\documentclass{article}

\usepackage[nonatbib,final]{nips_2016}

\usepackage[utf8]{inputenc} 
\usepackage[T1]{fontenc}    
\usepackage{hyperref}       
\usepackage{url}            
\usepackage{booktabs}       
\usepackage{amsfonts}       
\usepackage{nicefrac}       
\usepackage{microtype}      
\usepackage[nottoc]{tocbibind}
\usepackage{textcomp}
\usepackage[square,numbers, sort]{natbib}
\usepackage{graphicx}
\usepackage{amssymb}
\usepackage{graphicx}
\usepackage{subfigure}
\usepackage{threeparttable}
\usepackage[marginal, flushmargin]{footmisc}
\usepackage{algorithm}
\usepackage{algorithmic}
\usepackage{multirow}
\usepackage{bm}
\usepackage{amsmath}

\usepackage{comment}
\usepackage{color}
\usepackage{etoolbox}
\newbool{inccomment}
\setbool{inccomment}{true}
\newcommand{\hl}[1]{\ifbool{inccomment}{{\color{magenta}#1}}{}}
\newcommand{\ww}[1]{\ifbool{inccomment}{{\color{blue} #1}}{}}
\newcommand{\yc}[1]{\ifbool{inccomment}{{\color{red} #1}}{}}

\title{Learning Structured Sparsity in Deep Neural Networks}

\author{
  Wei Wen \\
  University of Pittsburgh\\
  \texttt{wew57@pitt.edu} \\
  \And
  Chunpeng Wu \\
  University of Pittsburgh\\
  \texttt{chw127@pitt.edu} \\
  \And
  Yandan Wang \\
  University of Pittsburgh\\
  \texttt{yaw46@pitt.edu} \\
  \And
  Yiran Chen \\
  University of Pittsburgh\\
  \texttt{yic52@pitt.edu} \\
  \And
  Hai Li \\
  University of Pittsburgh\\
  \texttt{hal66@pitt.edu} \\
}

\begin{document}

\maketitle

\begin{abstract}
  High demand for computation resources severely hinders deployment of large-scale Deep Neural Networks (DNN) in resource constrained devices. In this work, we propose a \textit{Structured Sparsity Learning} (SSL) method to regularize the structures (\textit{i.e.}, filters, channels, filter shapes, and layer depth) of DNNs. 
  SSL can: (1) learn a compact structure from a bigger DNN to reduce computation cost; (2) obtain a hardware-friendly structured sparsity of DNN to efficiently accelerate the DNN's evaluation. 
  Experimental results show that SSL achieves on average 5.1$\times$ and 3.1$\times$ speedups of convolutional layer computation of \textit{AlexNet} against CPU and GPU, respectively, with off-the-shelf libraries. These speedups are about twice speedups of non-structured sparsity; 
  (3) regularize the DNN structure to improve classification accuracy. The results show that for CIFAR-10, regularization on layer depth can reduce 20 layers of a Deep Residual Network (\textit{ResNet}) to 18 layers while improve the accuracy from 91.25\% to 92.60\%, which is still slightly higher than that of original \textit{ResNet} with 32 layers. For \textit{AlexNet}, structure regularization by SSL also reduces the error by $\sim{1\%}$. Our source code can be found at \url{https://github.com/wenwei202/caffe/tree/scnn}
  
\end{abstract}

\section{Introduction}
\label{sec:intro}

\begin{figure}
	\centering
	\includegraphics[width=0.8\textwidth,natwidth=1,natheight=1]{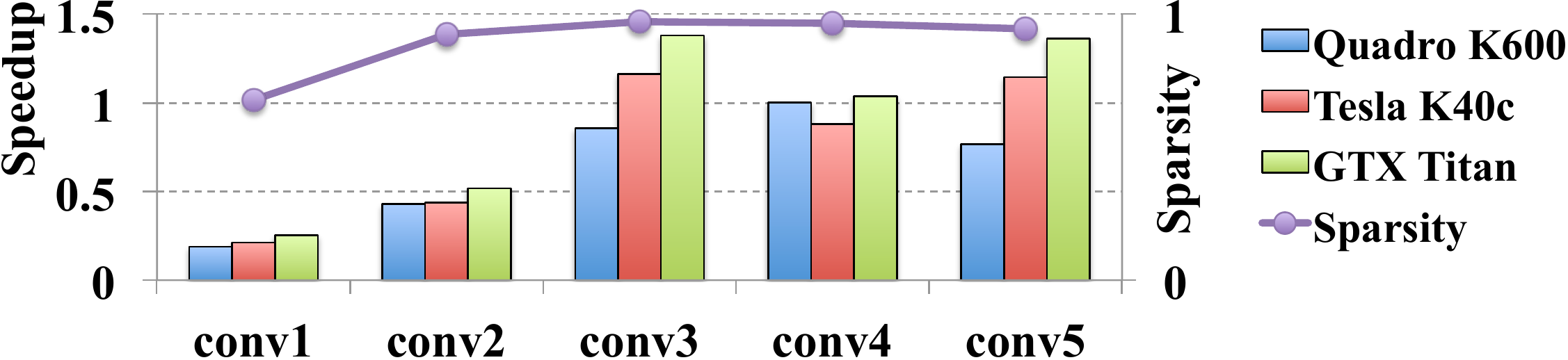}
	\vspace{-6pt}
	\caption{Evaluation speedups of AlexNet on GPU platforms and the sparsity. conv1 refers to convolutional layer 1, and so forth. Baseline is profiled by GEMM of cuBLAS. The sparse matrixes are stored in the format of Compressed Sparse Row (CSR) and accelerated by cuSPARSE.}
	\label{gpu_speedup_vs_sparsity}
\end{figure}

Deep neural networks (DNN), especially deep convolutional neural networks (CNN), made remarkable success in visual tasks\cite{Alex_NIPS2012_4824}\cite{RCNN_2014_CVPR}\cite{Vggnet_2014}\cite{GoogleNet_2015}\cite{Kaiming_ResNet_ICCV} by leveraging large-scale networks learning from a huge volume of data.
Deployment of such big models, however, is computation-intensive and memory-intensive.
To reduce computation cost, many studies are performed to compress the scale of DNN, including sparsity regularization\cite{Liu_CVPR2015}, connection pruning\cite{Han_NIPS2015}\cite{Han_ICLR2016} and low rank approximation \cite{Denil_NIPS2013_5025}\cite{Denton_NIPS2014}\cite{Max_J_arxiv2014}\cite{Yani_arxiv_2015}\cite{Cheng_arxiv_2015}.
Sparsity regularization and connection pruning approaches, however, often produce non-structured random connectivity in DNN
and thus, irregular memory access that adversely impacts \emph{practical} acceleration in hardware platforms.
Figure \ref{gpu_speedup_vs_sparsity} depicts practical speedup of each layer of a \textit{AlexNet}, which is non-structurally sparsified by $\ell_1$-norm.
Compared to original model, the accuracy loss of the sparsified model is controlled within 2\%.
Because of the poor data locality associated with the scattered weight distribution, the achieved speedups are either very limited or negative even the actual sparsity is high, say, >95\%. We define sparsity as the ratio of zeros in this paper.
In recently proposed low rank approximation approaches, the DNN is trained first and then each trained weight tensor is decomposed and approximated by a product of smaller factors. Finally, fine-tuning is performed to restore the model accuracy.
Low rank approximation is able to achieve practical speedups because it coordinates model parameters in dense matrixes and avoids the locality problem of non-structured sparsity regularization.
However, low rank approximation can only obtain the compact structure within each layer, and the structures of the layers are fixed during fine-tuning such that costly reiterations of decomposing and fine-tuning are required to find an optimal weight approximation for performance speedup and accuracy retaining.

Inspired by the facts that 
(1) there is redundancy across filters and channels~\cite{Max_J_arxiv2014}; 
(2) shapes of filters are usually fixed as cuboid but enabling arbitrary shapes can potentially eliminate unnecessary computation imposed by this fixation; 
and (3) depth of the network is critical for classification but deeper layers cannot always guarantee a lower error because of the exploding gradients and degradation problem~\cite{Kaiming_ResNet_ICCV}, 
we propose Structured Sparsity Learning (SSL) method to \emph{directly} learn a compressed structure of deep CNNs by group Lasso regularization during the training. SSL is a generic regularization to adaptively adjust mutiple structures in DNN, including structures of filters, channels, and filter shapes within each layer, and structure of depth beyond the layers. 
SSL combines structure regularization (on DNN for classification accuracy) with locality optimization (on memory access for computation efficiency), offering not only well-regularized big models with improved accuracy but greatly accelerated computation (\textit{e.g.} 5.1$\times$ on CPU and 3.1$\times$ on GPU for \textit{AlexNet}). 
\section{Related works}
\label{sec:related}
\vspace{-6pt}

\textbf{\textit{Connection pruning and weight sparsifying.~}}
Han \textit{et al.} \cite{Han_NIPS2015}\cite{Han_ICLR2016} reduced number of parameters of \textit{AlexNet} by 9$\times$ and \textit{VGG-16} by 13$\times$ using connection pruning. Since most reduction is achieved on fully-connected layers, the authors obtained 3$\times$ to 4$\times$ layer-wise speedup for fully-connected layers. However, no practical speedups of convolutional layers are observed because of the issue shown in Figure \ref{gpu_speedup_vs_sparsity}. As convolution is the computational bottleneck and many new DNNs use fewer fully-connected layers, \textit{e.g.}, only 3.99\% parameters of \textit{ResNet-152} in \cite{Kaiming_ResNet_ICCV} are from fully-connected layers, compression and acceleration on convolutional layers become essential. Liu \textit{et al.} \cite{Liu_CVPR2015} achieved >90\% sparsity of convolutional layers in \textit{AlexNet} with 2\% accuracy loss, and bypassed the issue shown in Figure \ref{gpu_speedup_vs_sparsity} by hardcoding the sparse weights into program, achieving layer-wise 4.59$\times$ speedup on a CPU. In this work, we also focus on convolutional layers. Compared to the above techniques, our SSL method can coordinate sparse weights in adjacent memory space and achieve higher speedups with the same accuracy. Note that hardware and program optimizations can further boost the system performance on top of the level of SSL but are not covered in this work.

\textbf{\textit{Low rank approximation.~}}
Denil \textit{et al.} \cite{Denil_NIPS2013_5025} predicted 95\% parameters in a DNN by exploiting the redundancy across filters and channels. 
Inspired by it, Jaderberg \textit{et al.} \cite{Max_J_arxiv2014} achieved 4.5$\times$ speedup on CPUs for scene text character recognition and Denton \textit{et al.} \cite{Denton_NIPS2014} achieved 2$\times$ speedups on both CPUs and GPUs for the first two layers. Both of the works used \textit{Low Rank Approximation} (LRA) with $\sim$1\% accuracy drop. 
\cite{Cheng_arxiv_2015}\cite{Yani_arxiv_2015} improved and extended LRA to larger DNNs.
However, the network structure compressed by LRA is fixed; reiterations of decomposing, training/fine-tuning, and cross-validating are still needed to find an optimal structure for accuracy and speed trade-off.
As number of hyper-parameters in LRA method increases linearly with layer depth \cite{Denton_NIPS2014}\cite{Cheng_arxiv_2015}, the search space increases linearly or even polynomially for very deep DNNs. 
Comparing to LRA, our contributions are: 
(1) SSL can dynamically optimize the compactness of DNN structure with only one hyper-parameter and no reiterations; 
(2) besides the redundancy within the layers, SSL also exploits the necessity of deep layers and reduce them; 
(3) DNN filters regularized by SSL have \emph{lower} rank approximation, so it can work together with LRA for more efficient model compression.

\textbf{\textit{Model structure learning.~}}
Group Lasso \cite{GroupLasso_2006} is an efficient regularization to learn sparse structures. 
Kim \textit{et al.} \cite{kim2010tree} used group Lasso to regularize the structure of correlation tree for multi-task regression problem and reduced prediction errors. Liu \textit{et al.} \cite{Liu_CVPR2015} utilized group Lasso to constrain the scale of the structure of LRA.
To adapt DNN structure to different databases, Feng \textit{et al.} \cite{Feng_2015_ICCV} learned the appropriate number of filters in DNN. 
Different from these prior arts, we apply group Lasso to regularize multiple DNN structures (filters, channels, filter shapes, and layer depth). Our source code can be found at \url{https://github.com/wenwei202/caffe/tree/scnn}.

\section{Structured Sparsity Learning Method for DNNs}
\label{sec:method}

We focus mainly on the \textit{Structured Sparsity Learning} (SSL) on convolutional layers to regularize the structure of DNNs. We first propose a generic method to regularize structures of DNN in Section \ref{sec:method:generic}, and then specify the method to structures of filters, channels, filter shapes and depth in section \ref{sec:method:specific}. Variants of formulations are also discussed from computational efficiency viewpoint in Section \ref{sec:method:computation}.

\begin{figure}
\centering
\includegraphics[width=0.9\textwidth]{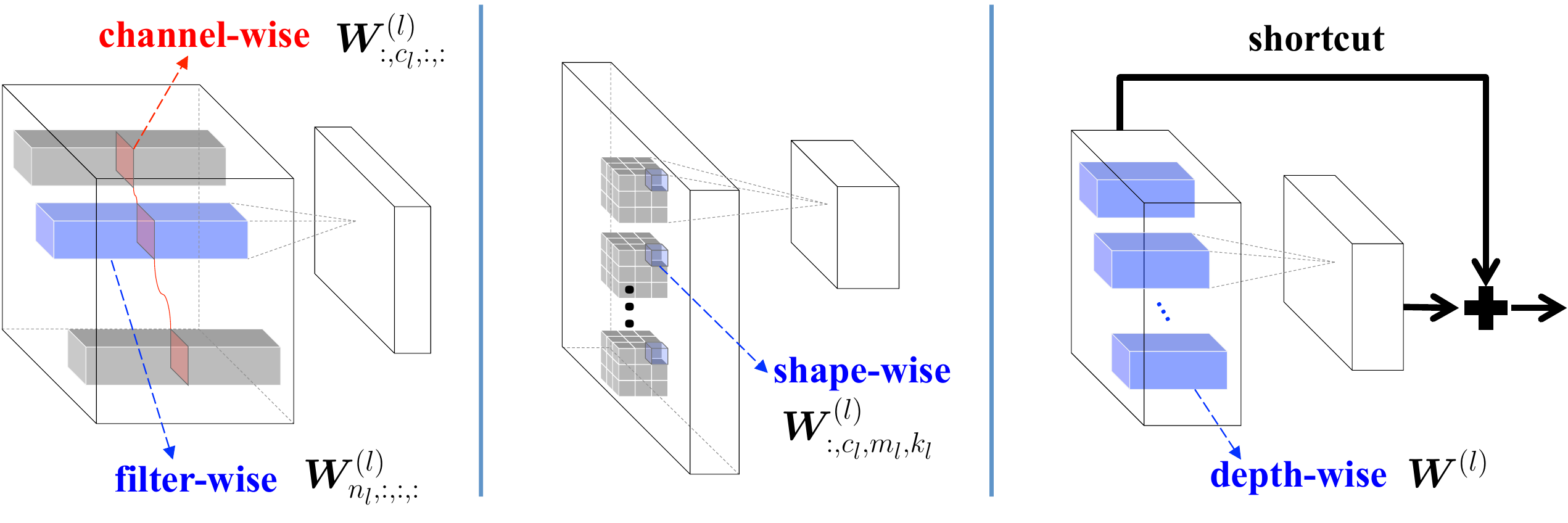}
\vspace{-9pt}
\caption{The proposed structured sparsity learning (SSL) for DNNs. Weights in filters are split into multiple groups. 
Through group Lasso regularization, a more compact DNN is obtained by removing some groups.
The figure illustrates the filter-wise, channel-wise, shape-wise, and depth-wise structured sparsity that were explored in the work.
}
\vspace{-6pt}
\label{fig:ssl}
\end{figure}

\subsection{Proposed structured sparsity learning for generic structures}
\label{sec:method:generic}
Suppose weights of convolutional layers in a DNN form a sequence of 4-D tensors $\bm{W}^{(l)}~\in~\mathbb{R}^{N_l \times C_l \times M_l \times K_l}$,
where $N_l$, $C_l$, $M_l$ and $K_l$ are the dimensions of the $l$\textit{-th} ($1\leq l \leq L$) weight tensor along the axes of filter, channel, spatial height and spatial width, respectively.
$L$ denotes the number of convolutional layers.
Then the proposed generic optimization target of a DNN with structured sparsity regularization can be formulated as:
\begin{equation}
\small
E(\bm{W}) = E_D(\bm{W}) + \lambda\cdot R(\bm{W})+\lambda_g\cdot\sum\limits_{l=1}^{L}R_g \left(\bm{W}^{(l)}\right).
\label{eq:generic_opt}
\end{equation}
Here $\bm{W}$ represents the collection of all weights in the DNN;
$E_D(\bm{W})$ is the loss on data;
$R(\cdot)$ is non-structured regularization applying on every weight, \textit{e.g.}, $\ell_2$-norm;
and $R_g(\cdot)$ is the structured sparsity regularization on each layer.
Because \textit{Group Lasso} can effectively zero out all weights in some groups \cite{GroupLasso_2006}\cite{kim2010tree}, we adopt it in our SSL.
The regularization of group Lasso on a set of weights $\bm{w}$ can be represented as $R_g(\bm{w})=\sum_{g=1}^{G}||\bm{w}^{(g)}||_g$,
where $\bm{w}^{(g)}$ is a group of partial weights in $\bm{w}$ and $G$ is the total number of groups. Different groups may overlap.
Here $||\cdot||_g$ is the group Lasso, or $||\bm{w}^{(g)}||_g = \sqrt{\sum\nolimits_{i=1}^{|\bm{w}^{(g)}|} \left( w_i^{(g)}\right)^2 }$, where $|\bm{w}^{(g)}|$ is the number of weights in $\bm{w}^{(g)}$.

\subsection{Structured sparsity learning for structures of filters, channels, filter shapes and depth}
\label{sec:method:specific}
In SSL, the learned ``structure'' is decided by the way of splitting groups of $\bm{w}^{(g)}$.
We investigate and formulate the \textit{filer-wise}, \textit{channel-wise}, \textit{shape-wise}, and \textit{depth-wise} structured sparsity in Figure \ref{fig:ssl}.
For simplicity, the $R(\cdot)$ term of Eq.~(\ref{eq:generic_opt}) is omitted in the following formulation expressions. 

\textbf{\textit{Penalizing unimportant filers and channels.} }
Suppose $\bm{W}_{n_l,:,:,:}^{(l)}$ is the $n_l$\textit{-th} filter and $\bm{W}_{:,c_l,:,:}^{(l)}$ is the $c_l$\textit{-th} channel of all filters in the $l$\textit{-th} layer.
The optimization target of learning the filter-wise and channel-wise structured sparsity can be defined as
\begin{equation}
\small
E(\bm{W}) = E_D(\bm{W})
 + \lambda_n \cdot \sum_{l=1}^{L}\left( \sum_{n_l=1}^{N_l}||\bm{W}^{(l)}_{n_l,:,:,:}||_g  \right)
 + \lambda_c \cdot \sum_{l=1}^{L}\left( \sum_{c_l=1}^{C_l}||\bm{W}^{(l)}_{:,c_l,:,:}||_g  \right).
\label{eq:target_filter_channel}
\end{equation}
As indicated in Eq.~(\ref{eq:target_filter_channel}), our approach tends to remove less important filters and channels.
Note that zeroing out a filter in the $l$\textit{-th} layer results in a dummy zero output feature map, which in turn makes 
a corresponding channel in the $(l+1)$\textit{-th} layer useless.
Hence, we combine the filter-wise and channel-wise structured sparsity in the learning simultaneously.

\textbf{\textit{Learning arbitrary shapes of filers. }}
As illustrated in Figure~\ref{fig:ssl}, $\bm{W}_{:,c_l,m_l,k_l}^{(l)}$ denotes the vector of all corresponding weights located at spatial position of $(m_l,k_l)$ in the 2D filters across the $c_l$\textit{-th} channel.
Thus, we define $\bm{W}_{:,c_l,m_l,k_l}^{(l)}$ as the \textit{shape fiber} related to learning arbitrary filter shape because a homogeneous non-cubic filter shape can be learned by zeroing out some shape fibers.
The optimization target of learning shapes of filers becomes:
\begin{equation}
\small
E(\bm{W}) = E_D(\bm{W})
+ \lambda_s \cdot \sum_{l=1}^{L}\left( \sum_{c_l=1}^{C_l}\sum_{m_l=1}^{M_l}\sum_{k_l=1}^{K_l}||\bm{W}^{(l)}_{:,c_l,m_l,k_l}||_g  \right).
\label{eq:target_shape}
\end{equation}

\textbf{\textit{Regularizing layer depth. }}
We also explore the depth-wise sparsity to regularize the depth of DNNs in order to improve accuracy and reduce computation cost.
The corresponding optimization target is $E(\bm{W})~=~E_D(\bm{W}) + \lambda_d \cdot \sum_{l=1}^{L}||\bm{W}^{(l)}||_g$.
Different from other discussed sparsification techniques, zeroing out all the filters in a layer will cut off the message propagation in the DNN so that the output neurons cannot perform any classification.
Inspired by the structure of highway networks \cite{HighwayNet_2015} and deep residual networks \cite{Kaiming_ResNet_ICCV}, we propose to leverage the shortcuts across layers to solve this issue.
As illustrated in Figure~\ref{fig:ssl}, even when SSL removes an entire unimportant layers, feature maps will still be forwarded through the shortcut.

\subsection{Structured sparsity learning for computationally efficient structures}
\label{sec:method:computation}
All proposed schemes in section \ref{sec:method:specific} can learn a compact DNN for computation cost reduction.
Moreover, some variants of the formulations of these schemes can directly learn structures that can be efficiently computed.

\textbf{\textit{2D-filter-wise sparsity for convolution.} }
3D convolution in DNNs essentially is a composition of 2D convolutions.
To perform efficient convolution, we explored a fine-grain variant of filter-wise sparsity, namely, \textit{2D-filter-wise} sparsity, to spatially enforce group Lasso on each 2D filter of $\bm{W}^{(l)}_{n_l,c_l,:,:}$.
The saved convolution is proportional to the percentage of the removed 2D filters. 
The fine-grain version of filter-wise sparsity can more efficiently reduce the computation associated with convolution: Because the group sizes are much smaller and thus the weight updating gradients are shaper, it helps group Lasso to quickly obtain a high ratio of zero groups for a large-scale DNN.

\textbf{\textit{Combination of filter-wise and shape-wise sparsity for GEMM.} }
Convolutional computation in DNNs is commonly converted to modality of \textit{GEneral Matrix Multiplication} (GEMM) by lowering weight tensors and feature tensors to matrices \cite{cuDNN_arxiv2014}.
For example, in Caffe~\cite{Caffe_2014}, a 3D filter $\bm{W}^{(l)}_{n_l,:,:,:}$ is reshaped to a row in the weight matrix where each column is the collection of weights $\bm{W}^{(l)}_{:,c_l,m_l,k_l}$ related to shape-wise sparsity.
Combining filter-wise and shape-wise sparsity can directly reduce the dimension of weight matrix in GEMM by removing zero rows and columns. In this context, we use \textit{row-wise} and \textit{column-wise} sparsity as the interchangeable terminology of \textit{filter-wise} and \textit{shape-wise sparsity}, respectively.

\section{Experiments}
\label{sec:exp}

We evaluated the effectiveness of our SSL using published models on three databases -- MNIST, CIFAR-10, and ImageNet.
Without explicit explanation, SSL starts with the network whose weights are initialized by the baseline, and speedups are measured in matrix-matrix multiplication by Caffe in a single-thread Intel Xeon E5-2630 CPU .

\subsection{\textit{LeNet} and multilayer perceptron on MNIST}
\label{sec:exp:minist}

In the experiment of MNIST, we examined the effectiveness of SSL in two types of networks:
\textit{LeNet}~\cite{lecun1998gradient} implemented by Caffe 
and a \textit{multilayer perceptron} (\textit{MLP}) network. 
Both networks were trained without data augmentation.

\begin{table}[t]
  \caption{Results after penalizing unimportant filters and channels in \textit{LeNet}}
  \label{tab:fc:lenet}
  \centering
  \small
  \vspace{-6pt}
  	\begin{tabular}{ccccccc}
  		\toprule
  		\textit{LeNet} \# & Error & Filter \# \textsuperscript{\textsection} & Channel \# \textsuperscript{\textsection} & FLOP \textsuperscript{\textsection} & Speedup \textsuperscript{\textsection} \\
  		\midrule
  		
  		1 (\textit{baseline}) & 0.9\% & 20---50 & 1---20 & 100\%---100\% & 1.00$\times$---1.00$\times$ \\
  		
  		2 & 0.8\% & 5---19 & 1---4 & 25\%---7.6\% & 1.64$\times$---5.23$\times$ \\
  		
  		3 & 1.0\% & 3---12 & 1---3 & 15\%---3.6\% & 1.99$\times$---7.44$\times$ \\
  		\bottomrule
  		\multicolumn{6}{l}{\textsuperscript{\textsection}In the order of \textit{conv1}---\textit{conv2}}
  	\end{tabular}
\end{table}

\textbf{\textit{LeNet}:}
When applying SSL to \textit{LeNet}, we constrain the network with filter-wise and channel-wise sparsity in convolutional layers to penalize unimportant filters and channels.
Table~\ref{tab:fc:lenet} summarizes the remained filters and channels, \textit{floating-point operations} (FLOP), and practical speedups.
In the table, \textit{LeNet~1} is the baseline and the others are the results after applying SSL in different strengths of structured sparsity regularization.
The results show that our method achieves the similar error ($\pm0.1\%$) with much fewer filters and channels, and saves significant FLOP and computation time.

\begin{figure}[htp]
	\centering
	\vspace{-6pt}
	\subfigure
	{
		\includegraphics[width=1.0\textwidth]{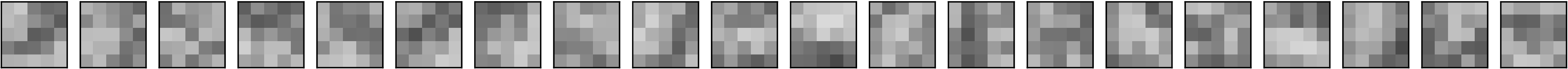}
	}
	\vfill
	\vspace{-10pt}
	\subfigure
	{
		\includegraphics[width=1.0\textwidth]{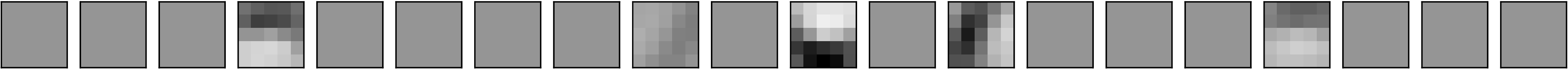}
	}
	\vfill
	\vspace{-10pt}
	\subfigure
	{
		\includegraphics[width=1.0\textwidth]{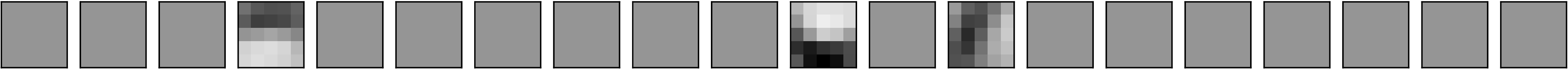}
	}
	\vspace{-15pt}
	\caption{Learned \textit{conv1} filters in \textit{LeNet 1} (top), \textit{LeNet 2} (middle) and \textit{LeNet 3} (bottom)}
	
	\label{fig:lenet_conv1_filters}
\end{figure}

To demonstrate the impact of SSL on the structures of filters, we present all learned \textit{conv1} filters in Figure~\ref{fig:lenet_conv1_filters}.
It can be seen that most filters in \textit{LeNet 2} are entirely zeroed out except for five most important detectors of stroke patterns that are sufficient for feature extraction.
The accuracy of \textit{LeNet~3} (that further removes the weakest and redundant stroke detector) drops only 0.2\% from that of \textit{LeNet~2}.
Compared to the random and blurry filter patterns in \textit{LeNet 1} that resulted from the high freedom of parameter space, the filters in \textit{LeNet 2 \& 3} are regularized and converge to smoother and more natural patterns.
This explains why our proposed SSL obtains the same-level accuracy but has much less filters.
The smoothness of the filters are also observed in the deeper layers.
\begin{table}[t]
  \caption{Results after learning filter shapes in \textit{LeNet}}
  \label{tab:shape:lenet}
  \centering
  \small
  \vspace{-6pt}
  	\begin{tabular}{cccccc}
  		\toprule
  		\textit{LeNet} \#& Error & Filter size \textsuperscript{\textsection} & Channel \# & FLOP & Speedup \\
  		\midrule
  		
  		1~(\textit{baseline}) & 0.9\%  & 25---500 & 1---20 & 100\%---100\% & 1.00$\times$---1.00$\times$ \\
  		\cmidrule{1-6}
  		
  		4 & 0.8\%  & 21---41 & 1---2 & 8.4\%---8.2\% & 2.33$\times$---6.93$\times$ \\
  		\cmidrule{1-6}
  		
  		5 & 1.0\%  & 7---14 & 1---1 &  1.4\%---2.8\% & 5.19$\times$---10.82$\times$ \\
  		\bottomrule
  		\multicolumn{6}{l}{\textsuperscript{\textsection} The sizes of filters after removing zero shape fibers, in the order of \textit{conv1}---\textit{conv2}}
  	\end{tabular}
  	\vspace{-9pt}
\end{table}

The effectiveness of the shape-wise sparsity on \textit{LeNet} is summarized in Table~\ref{tab:shape:lenet}.
The baseline \textit{LeNet~1} has \textit{conv1} filters with a regular $5\times5$ square (size = 25) while \textit{LeNet 5} reduces the dimension that can be constrained by a $2\times4$ rectangle (size = 7).
The 3D shape of \textit{conv2} filters in the baseline is also regularized to the 2D shape in \textit{LeNet 5} within only one channel, indicating that only one filter in \textit{conv1} is needed. This fact significantly saves FLOP and computation time.

\textbf{\textit{MLP}:}
Besides convolutional layers, our proposed SSL can be extended to learn the structure (\textit{i.e.} the number of neurons) of fully-connected layers.
We enforce the group Lasso regularization on all the input (or output) connections of each neuron.
A neuron whose input connections are all zeroed out can degenerate to a bias neuron in the next layer;
similarly, a neuron can degenerate to a removable dummy neuron if all of its output connections are zeroed out.
\begin{figure}
	\centering
	\subfigure[]
	{
		\label{fig:mlp}
		\includegraphics[width=0.74\textwidth]{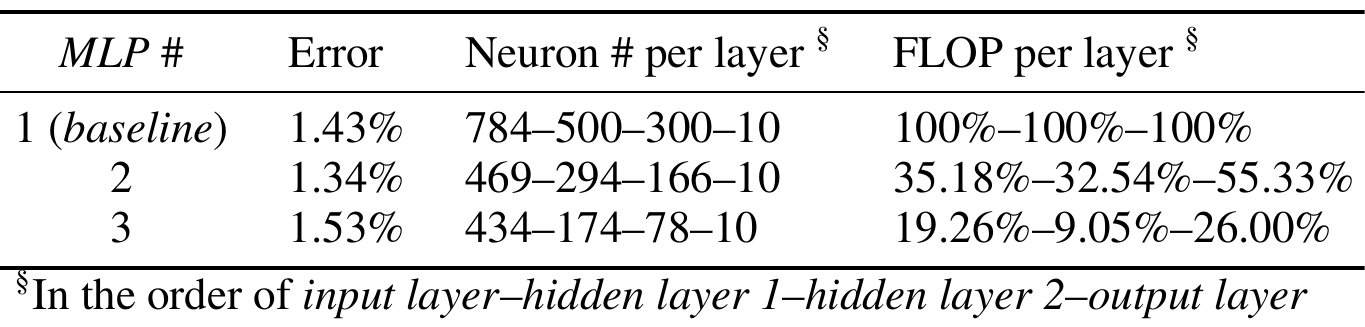}
	}
	\subfigure[]
	{
		\label{fig:fanout_mlp}
		\includegraphics[width=0.20\textwidth]{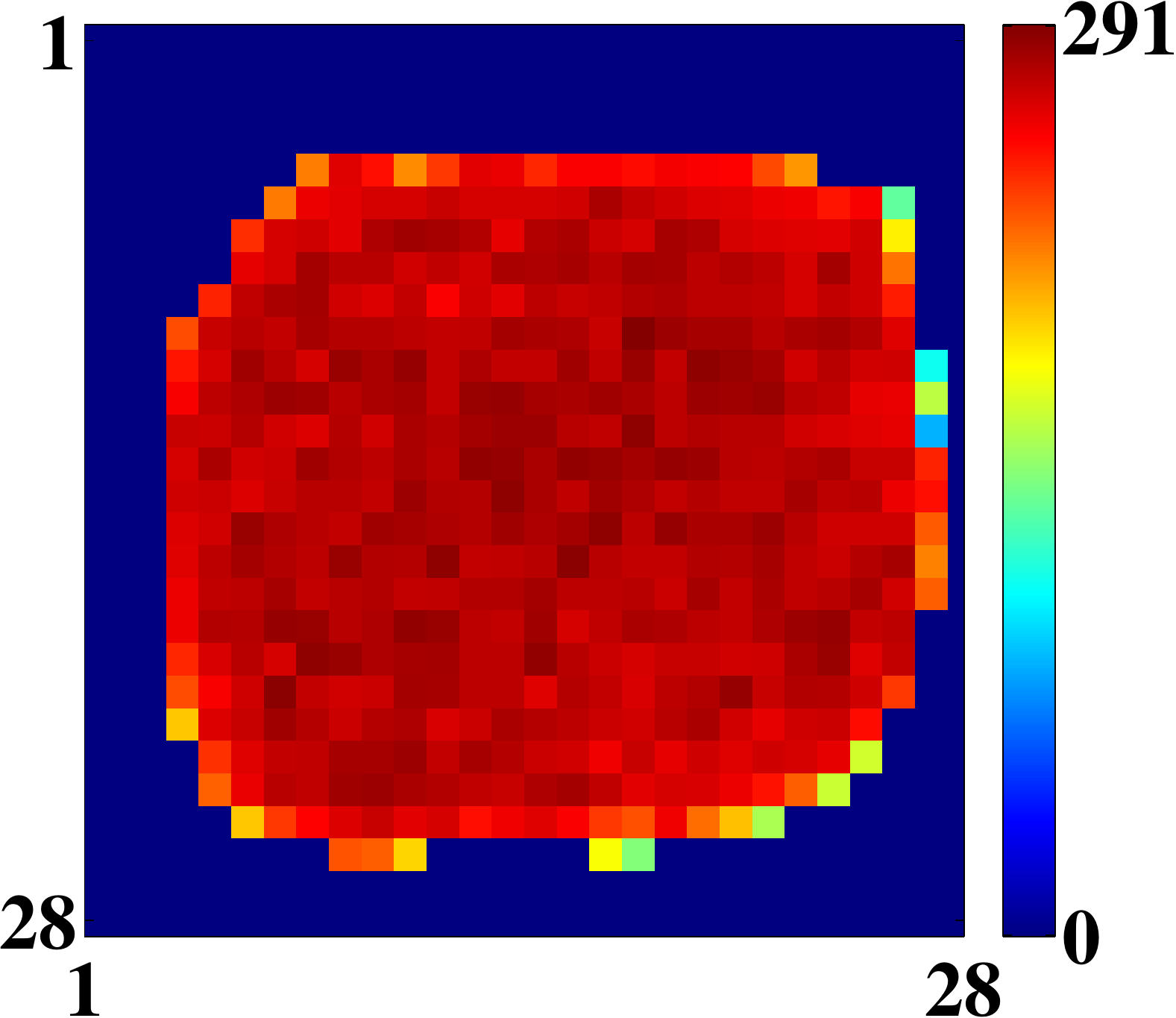}
	}
	\vspace{-8pt}
	\caption{(a) Results of learning the number of neurons in \textit{MLP}. (b) the connection numbers of input neurons (\textit{i.e.} pixels) in \textit{MLP 2} after SSL.}
\end{figure}
Figure~\ref{fig:mlp} summarizes the learned structure and FLOP of different \textit{MLP} networks.
The results show that SSL can not only remove hidden neurons but also discover the sparsity of images.
For example, Figure~\ref{fig:fanout_mlp} depicts the number of connections of each input neuron in \textit{MLP 2}, where 40.18\% of input neurons have zero connections and they concentrate at the boundary of the image.
Such a distribution is consistent with our intuition: handwriting digits are usually written in the center and pixels close to the boundary contain little discriminative classification information.

\subsection{\textit{ConvNet} and \textit{ResNet} on CIFAR-10}
\label{sec:exp:cifar}

We implemented the \textit{ConvNet} of \cite{Alex_NIPS2012_4824} and \textit{deep residual networks} (\textit{ResNet}) \cite{Kaiming_ResNet_ICCV} on CIFAR-10.
When regularizing filters, channels, and filter shapes, the results and observations of both networks are similar to that of the MNIST experiment.
Moreover, we simultaneously learn the filter-wise and shape-wise sparsity to reduce the dimension of weight matrix in GEMM of \textit{ConvNet}.
We also learn the depth-wise sparsity of \textit{ResNet} to regularize the depth of the DNNs.

\begin{figure}[htp]
	\centering
	\vspace{-6pt}
	\subfigure
	{
		\includegraphics[width=1.0\textwidth]{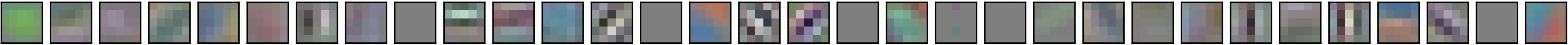}
	}
	\vfill
	\vspace{-10pt}
	\subfigure
	{
		\includegraphics[width=1.0\textwidth]{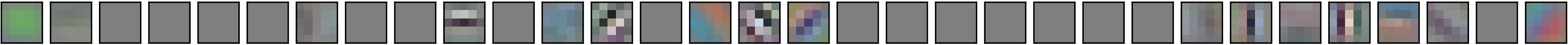}
	}
	\vfill
	\vspace{-10pt}
	\subfigure
	{
		\includegraphics[width=1.0\textwidth]{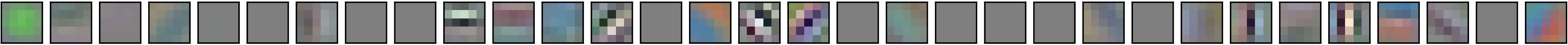}
	}
	\vspace{-15pt}
	\caption{Learned \textit{conv1} filters in \textit{ConvNet~1} (top), \textit{ConvNet~2} (middle) and \textit{ConvNet~3} (bottom)}
	
	\label{fig:cifar10_conv1_filters}
	\vspace{-6pt}
\end{figure}

\textbf{\textit{ConvNet}:}
We use the network from Alex Krizhevsky \textit{et al.} \cite{Alex_NIPS2012_4824} as the baseline and implement it using Caffe.
All the configurations remain the same as the original implementation except that we added a dropout layer with a ratio of 0.5 in the fully-connected layer to avoid over-fitting. 
\textit{ConvNet} is trained without data augmentation.
Table~\ref{tab:convnet:cifar10} summarizes the results of three \textit{ConvNet} networks. 
Here, the row/column sparsity of a weight matrix is defined as the percentage of all-zero rows/columns. 
Figure~\ref{fig:cifar10_conv1_filters} shows their learned \textit{conv1} filters.
In Table~\ref{tab:convnet:cifar10}, SSL can reduce the size of weight matrix in \textit{ConvNet 2} by 50\%, 70.7\% and 36.1\% for each convolutional layer and achieve good speedups without accuracy drop.
Surprisingly, without SSL, four \textit{conv1} filters of the baseline are actually all-zeros as shown in Figure~\ref{fig:cifar10_conv1_filters}, 
demonstrating the great potential of filter sparsity.
When SSL is applied, half of \textit{conv1} filters in \textit{ConvNet 2} can be zeroed out without accuracy drop. 

\begin{table}[t]
  \caption{Learning row-wise and column-wise sparsity of \textit{ConvNet} on CIFAR-10 }
  \label{tab:convnet:cifar10}
  \centering
  \small
  \vspace{-6pt}
  	\begin{tabular}{cllll}
  		\toprule
  		\textit{ConvNet \#} & Error & Row sparsity \textsuperscript{\textsection} & Column sparsity \textsuperscript{\textsection} & Speedup \textsuperscript{\textsection} \\
  		\midrule
  		1 (\textit{baseline})& 17.9\%
  		& 12.5\%\textendash0\%\textendash0\%
  		& 0\%\textendash0\%\textendash0\%
  		& 1.00$\times$\textendash1.00$\times$\textendash1.00$\times$ \\
  		
  		2 & 17.9\%
  		& 50.0\%\textendash28.1\%\textendash1.6\%
  		& 0\%\textendash59.3\%\textendash35.1\%
  		& 1.43$\times$\textendash3.05$\times$\textendash1.57$\times$ \\
  		
  		3 & 16.9\%
  		& 31.3\%\textendash0\%\textendash1.6\%
  		& 0\%\textendash42.8\%\textendash9.8\%
  		& 1.25$\times$\textendash2.01$\times$\textendash1.18$\times$ \\
  		\bottomrule
  		\multicolumn{5}{l}{\textsuperscript{\textsection}in the order of \textit{conv1}\textendash\textit{conv2}\textendash\textit{conv3}}
  	\end{tabular}
  	\vspace{-6pt}
\end{table} 

On the other hand, in \textit{ConvNet 3}, SSL achieves 1.0\% ($\pm$0.16\%) lower error with a model even smaller than the baseline. 
In this scenario, SSL performs as a structure regularization to dynamically learn a better network structure (including the number of filters and filer shapes) to reduce the error.

\textbf{\textit{ResNet}:}
To investigate the necessary depth of DNNs required by SSL, we use a 20-layer deep residual networks (\textit{ResNet-20}) proposed in \cite{Kaiming_ResNet_ICCV} as the baseline. 
The network has 19 convolutional layers and 1 fully-connected layer. 
\textit{Identity shortcuts} are utilized to connect the feature maps with the same dimension 
while 1$\times$1 convolutional layers are chosen as shortcuts between the feature maps with different dimensions.
Batch normalization \cite{batchnorm_2015} is adopted after convolution and before activation. 
We use the same data augmentation and training hyper-parameters as that in \cite{Kaiming_ResNet_ICCV}. 
The final error of baseline is 8.82\%.
In SSL, the depth of \textit{ResNet-20} is regularized by depth-wise sparsity. 
Group Lasso regularization is only enforced on the convolutional layers between each pair of shortcut endpoints, 
excluding the first convolutional layer and all convolutional shortcuts.
After SSL converges, layers with all zero weights are removed and the net is finally fine-tuned with a base learning rate of 0.01, which is lower than that (i.e., 0.1) in the baseline.
\begin{figure}
	\centering
	\includegraphics[width=1.0\textwidth]{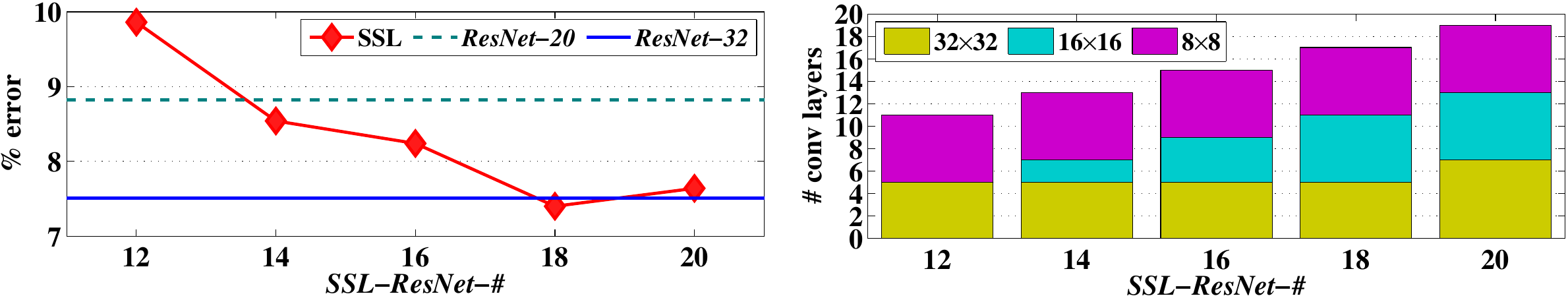}
	\vspace{-4pt}
	\caption{Error vs. layer number after depth regularization by SSL. \textit{ResNet-\#} is the original \textit{ResNet} in \cite{Kaiming_ResNet_ICCV} with \# layers. 
\textit{SSL-ResNet-\#} is the depth-regularized \textit{ResNet} by SSL with \# layers, including the last fully-connected layer.
32$\times$32 indicates the convolutional layers with an output map size of 32$\times$32, and so forth.}
	\label{fig:error_vs_depth}
	\vspace{-12pt}
\end{figure}

Figure~\ref{fig:error_vs_depth} plots the trend of the error vs. the number of layers under different strengths of depth regularizations. 
Compared with original \textit{ResNet} in \cite{Kaiming_ResNet_ICCV}, SSL learns a \textit{ResNet} with 14 layers (\textit{SSL-ResNet-14}) that reaching a lower error than the one of the baseline with 20 layers (\textit{ResNet-20}); 
\textit{SSL-ResNet-18} and \textit{ResNet-32} achieve an error of 7.40\% and 7.51\%, respectively. 
This result implies that SSL can work as a depth regularization to improve classification accuracy. 
Note that SSL can efficiently learn shallower DNNs without accuracy loss to reduce computation cost; 
however, it does not mean the depth of the network is not important. 
The trend in Figure~\ref{fig:error_vs_depth} shows that the test error generally declines as more layers are preserved. 
A slight error rise of \textit{SSL-ResNet-20} from \textit{SSL-ResNet-18} shows the suboptimal selection of the depth in the group of ``32$\times$32''.

\subsection{\textit{AlexNet} on ImageNet}
\label{sec:exp:alexnet}

To show the generalization of our method to large scale DNNs, we evaluate SSL using \textit{AlexNet} with ILSVRC 2012.
\textit{CaffeNet}~\cite{Caffe_2014} -- the replication of \textit{AlexNet}~\cite{Alex_NIPS2012_4824} with mirror changes, is used in our experiment.
All training images are rescaled to the size of 256$\times$256.
A 227$\times$227 image is randomly cropped from each scaled image and mirrored for data augmentation and only the center crop is used for validation.
The final top-1 validation error is 42.63\%.
In SSL, \textit{AlexNet} is first trained with structure regularization; when it converges, zero groups are removed to obtain a DNN with the new structure; finally, the network is fine-tuned without SSL to regain the accuracy.


\begin{figure}[b]
	\centering
	\vspace{-12pt}
	\subfigure[]
	{
	\includegraphics[width=0.3\textwidth]{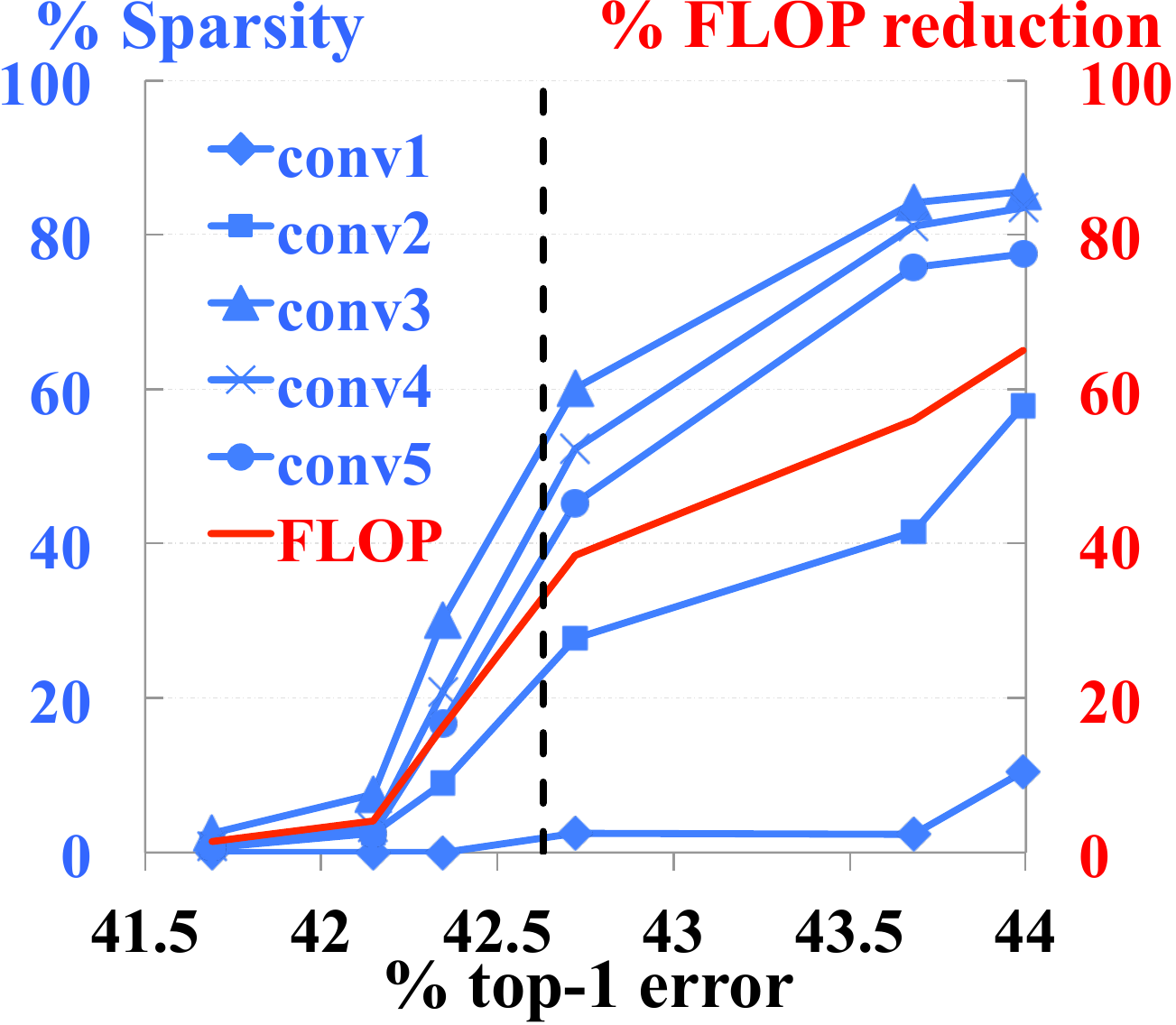}
	\label{fig:2dvsaccu}
	}
	\subfigure[]
	{
		\includegraphics[width=0.3\textwidth]{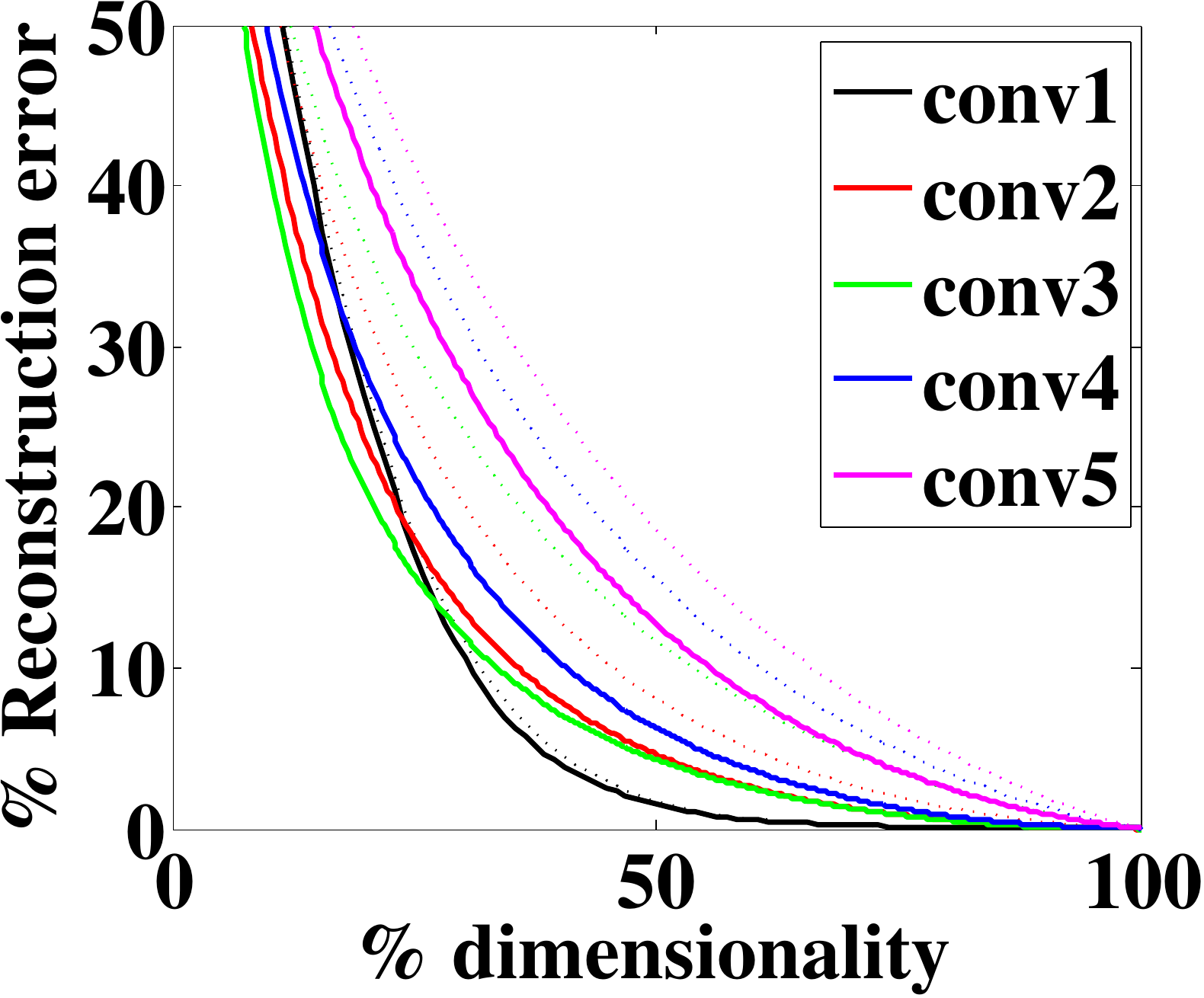}
		\label{fig:rankssl_alexnet}
	}
	\subfigure[]
	{
		\includegraphics[width=0.3\textwidth]{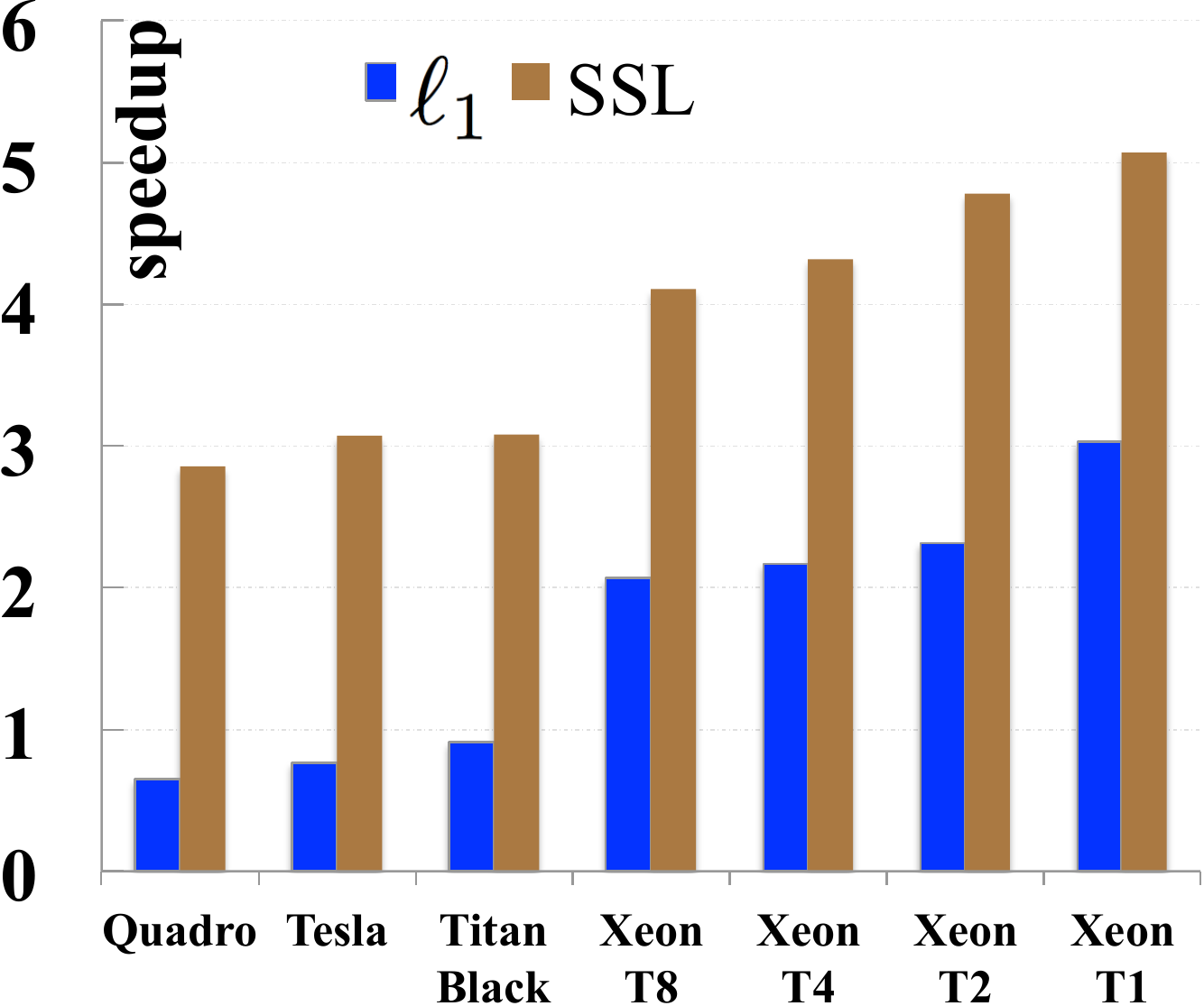}
		\label{fig:speed_scalability}
	}
	\vspace{-6pt}
	\caption{(a) 2D-filter-wise sparsity and FLOP reduction vs. top-1 error. Vertical dash line shows the error of original \textit{AlexNet}; 
(b) The reconstruction error of weight tensor vs. dimensionality. \textit{Principal Component Analysis} (PCA) is utilized to perform dimensionality reduction to exploit filter redundancy. The eigenvectors corresponding to the largest eigenvalues are selected as basis of lower-dimensional space. Dash lines denote the results of the baselines and solid lines indicate the ones of the \textit{AlexNet~5} in Table~\ref{tab:alexnet}; 
(c) Speedups of $\ell_1$-norm and SSL on various CPU and GPU platforms (In labels of x-axis, T\# is number of the maximum physical threads in Xeon CPU).
\textit{AlexNet~1} and \textit{AlexNet~2} in Table~\ref{tab:alexnet} are used as testbenches.}
	
\end{figure}

We first studied 2D-filter-wise and shape-wise sparsity by exploring the trade-offs between computation complexity and classification accuracy.
Figure~\ref{fig:2dvsaccu} shows the 2D-filter sparsity (the ratio between the removed 2D filters and total 2D filters) and the saved FLOP of 2D convolutions vs. the validation error. 
In Figure~\ref{fig:2dvsaccu}, deeper layers generally have higher sparsity as the group size shrinks and the number of 2D filters grows. 
2D-filter sparsity regularization can reduce the total FLOP by 30\%--40\% without accuracy loss or reduce the error of \textit{AlexNet} by $\sim$1\% down to 41.69\% by retaining the original number of parameters.
Shape-wise sparsity also obtains similar results -- 
In Table~\ref{tab:alexnet}, for example, \textit{AlexNet~5} achieves on average 1.4$\times$ layer-wise speedup on both CPU and GPU without accuracy loss after shape regularization;
The top-1 error can also be reduced down to 41.83\% if the parameters are retained.
In Figure~\ref{fig:2dvsaccu}, the obtained DNN with the lowest error has a very low sparsity, indicating that the number of parameters in a DNN is still important to maintain learning capacity.
In this case, SSL works as a regularization to add restriction of smoothness to the model in order to avoid over-fitting.
Figure~\ref{fig:rankssl_alexnet} compares the results of dimensionality reduction of weight tensors in the baseline and our SSL-regularized \textit{AlexNet}.
The results show that the smoothness restriction enforces parameter searching in lower-dimensional space and enables \textit{lower} rank approximation of the DNNs. 
Therefore, SSL can work together with low rank approximation to achieve even higher model compression.


Besides the above analyses, the computation efficiencies of structured sparsity and non-structured sparsity are compared in Caffe using standard off-the-shelf libraries, \textit{i.e.}, Intel Math Kernel Library on CPU and CUDA cuBLAS and cuSPARSE on GPU. 
We use SSL to learn a \textit{AlexNet} with high column-wise and row-wise sparsity as the representative of structured sparsity method.
$\ell_1$-norm is selected as the representative of non-structured sparsity method instead of connection pruning in \cite{Han_NIPS2015} because $\ell_1$-norm get a higher sparsity on convolutional layers as the results of \textit{AlexNet 3} and \textit{AlexNet 4} depicted in Table~\ref{tab:alexnet}.
Speedups achieved by SSL are measured by subroutines of GEMM where nonzero rows and columns in each weight matrix are concatenated in consecutive memory space. 
Note that compared to GEMM, the overhead of concatenation can be ignored. 
To measure the speedups of $\ell_1$-norm, sparse weight matrices are stored in the format of Compressed Sparse Row (CSR) and computed by sparse-dense matrix multiplication subroutines.
\begin{table}[b]
  \caption{Sparsity and speedup of \textit{AlexNet} on ILSVRC 2012}
  \label{tab:alexnet}
  \centering
  	\begin{tabular}{cccclllll}
  		\toprule[0.12 em]
  		\# & Method & Top1 err. & Statistics & conv1 & conv2 & conv3 & conv4 & conv5 \\

  		\midrule[0.12 em]
  		
  		\multirow{3}{*}{1} & \multirow{3}{*}{$\ell_1$} & \multirow{3}{*}{44.67\%} & 
  		sparsity  & 67.6\% & 92.4\% & 97.2\% & 96.6\% & 94.3\% \\
  		& & & 
  		CPU $\times$ & 0.80 & 2.91 & 4.84 & 3.83 & 2.76 \\
  		& & & 
  		GPU $\times$ & 0.25 & 0.52 & 1.38 & 1.04 & 1.36 \\
  		\cmidrule{1-9}

  		\multirow{4}{*}{2} & \multirow{4}{*}{SSL} & \multirow{4}{*}{44.66\%} &
  		column sparsity  & 0.0\% & 63.2\% & 76.9\% & 84.7\% & 80.7\% \\
  		& & & 
  		row sparsity  & 9.4\% & 12.9\% & 40.6\% & 46.9\% & 0.0\% \\
  		& & & 
  		CPU $\times$ & 1.05 & 3.37 & 6.27 & 9.73 & 4.93 \\
  		& & & 
  		GPU $\times$ & 1.00 & 2.37 & 4.94 & 4.03 & 3.05 \\
  		\cmidrule[0.12 em]{1-9}

  		3 & pruning\cite{Han_NIPS2015} & 42.80\% &
  		sparsity  & 16.0\% & 62.0\% & 65.0\% & 63.0\% & 63.0\% \\
  		\cmidrule{1-9}
  		
  		\multirow{3}{*}{4} & \multirow{3}{*}{$\ell_1$} & \multirow{3}{*}{42.51\%} &
  		sparsity  & 14.7\% & 76.2\% & 85.3\% & 81.5\% & 76.3\% \\
  		& & & 
  		CPU $\times$ & 0.34 & 0.99 & 1.30 & 1.10 & 0.93 \\
  		& & & 
  		GPU $\times$ & 0.08 & 0.17 & 0.42 & 0.30 & 0.32 \\
  		\cmidrule{1-9}
  		
  		\multirow{3}{*}{5} & \multirow{3}{*}{SSL} & \multirow{3}{*}{42.53\%} &
  		column sparsity  & 0.00\% & 20.9\% & 39.7\% & 39.7\% & 24.6\% \\
  		& & & 
  		CPU $\times$ & 1.00 & 1.27 & 1.64 & 1.68 & 1.32 \\
  		& & & 
  		GPU $\times$ & 1.00 & 1.25 & 1.63 & 1.72 & 1.36 \\
  		
  		\bottomrule[0.12 em]
  	\end{tabular}
\end{table}

Table~\ref{tab:alexnet} compares the obtained sparsity and speedups of $\ell_1$-norm and SSL on CPU (Intel Xeon) and GPU (GeForce GTX TITAN Black) under approximately the same errors, \textit{e.g.}, with acceptable or no accuracy loss.
For a fair comparison, after $\ell_1$-norm regularization, the DNN is also fine-tuned by disconnecting all zero-weighted connections so that 1.39\% accuracy is recovered for the \textit{AlexNet~1}. 
Our experiments show that the DNNs require a very high non-structured sparsity to achieve a reasonable speedup (The speedups are even negative when the sparsity is low).
SSL, however, can always achieve positive speedups.
With an acceptable accuracy loss, our SSL achieves on average 5.1$\times$ and 3.1$\times$ layer-wise acceleration on CPU and GPU, respectively.
Instead, $\ell_1$-norm achieves on average only 3.0$\times$ and 0.9$\times$ layer-wise acceleration on CPU and GPU, respectively. 
We note that at the same accuracy, our average speedup is indeed higher than that of \cite{Liu_CVPR2015} which adopts heavy hardware customization to overcome the negative impact of non-structured sparsity.
Figure~\ref{fig:speed_scalability} shows the speedups of $\ell_1$-norm and SSL on various platforms, including both GPU (Quadro, Tesla and Titan) and CPU (Intel Xeon E5-2630). 
SSL can achieve on average $\sim3\times$ speedup on GPU while non-structured sparsity obtain no speedup on GPU platforms. 
On CPU platforms, both methods can achieve good speedups and the benefit grows as the processors become weaker. 
Nonetheless, SSL can always achieve averagely $\sim2\times$ speedup compared to non-structured sparsity.

\section{Conclusion}
\label{sec:conclusion}
In this work, we have proposed a \textit{Structured Sparsity Learning} (SSL) method to regularize filter, channel, filter shape, and depth structures in deep neural networks (DNN). 
Our method can enforce the DNN to dynamically learn more compact structures without accuracy loss. 
The structured compactness of the DNN achieves significant speedups for the DNN evaluation both on CPU and GPU with off-the-shelf libraries. Moreover, a variant of SSL can be performed as structure regularization to improve classification accuracy of state-of-the-art DNNs. 

\subsubsection*{Acknowledgments}
This work was supported in part by NSF XPS-1337198 and NSF CCF-1615475. The authors thank Drs. Sheng Li and Jongsoo Park for valuable feedback on this work.

\small
\bibliography{nips_2016}

\end{document}